\documentclass[a4paper,fleqn]{cas-dc}

\usepackage{lineno,hyperref}
\usepackage{subfigure}
\usepackage{array} 
\usepackage{float}
\usepackage{soul} 
\usepackage{color, xcolor}
\usepackage{amsmath}
\usepackage{algorithm}
\usepackage{algorithmic}
\usepackage{graphicx}
\usepackage{sidecap}
\usepackage{tikz}
\usetikzlibrary{positioning} 
\usepackage{xcolor}
\usepackage{amsmath}
\definecolor{first}{RGB}{192,192,192}
\definecolor{second}{RGB}{137,149,152}
\definecolor{third}{RGB}{127,111,115}

\soulregister{\cite}7 
\soulregister{\citep}7 
\soulregister{\citet}7 
\soulregister{\ref}7 
\soulregister{\pageref}7 

\modulolinenumbers[5]

\usepackage[numbers]{natbib}

\def\tsc#1{\csdef{#1}{\textsc{\lowercase{#1}}\xspace}}
\tsc{WGM}
\tsc{QE}

\begin{document}
\let\WriteBookmarks\relax
\def\floatpagepagefraction{1}
\def\textpagefraction{.001}

\shorttitle{Understanding the Role of Pathways in a Deep Neural Network}    

\shortauthors{Lei Lyu et al.}  

\title [mode = title]{Understanding the Role of Pathways in a Deep Neural Network}

\author[1]{Lei Lyu}
\ead{lvlei@sdnu.edu.cn}

\author[1]{Chen Pang}
\ead{chenp0721@163.com}

\author[1]{Jihua Wang\corref{mycorrespondingauthor}}
\cortext[mycorrespondingauthor]{Corresponding author}
\ead{jihuaaw@126.com}

\address[1]{{School of Information Science and Engineering, Shandong Normal University},
	{Jinan},
	{China}}


\begin{abstract}
Deep neural networks have demonstrated superior performance in artificial intelligence applications, but the opaqueness of their inner working mechanism is one major drawback in their application. The prevailing unit-based interpretation is a statistical observation of stimulus-response data, which fails to show a detailed internal process of inherent mechanisms of neural networks. In this work, we analyze a convolutional neural network (CNN) trained in the classification task and present an algorithm to extract the diffusion pathways of individual pixels to identify the locations of pixels in an input image associated with object classes. The pathways allow us to test the causal components which are important for classification and the pathway-based representations are clearly distinguishable between categories. We find that the few largest pathways of an individual pixel from an image tend to cross the feature maps in each layer that is important for classification. And the large pathways of images of the same category are more consistent in their trends than those of different categories. We also apply the pathways to understanding adversarial attacks, object completion, and movement perception. Further, the total number of pathways on feature maps in all layers can clearly discriminate the original, deformed, and target samples.
\end{abstract}




\maketitle

\section{Introduction}

The interpretability of a deep neural network has become an important topic in practical applications, which can enable us to better understand the behavior of biological neural networks. What exactly is the inherent mechanism of data processing in deep networks? To explain this problem, some research has analyzed the semantics of individual units in a deep neural network \citep{a01}. However, these units can only make statistical observations of stimulus activation. Pathways of pixels can provide us with more insight than units when it comes to understanding the rationale behind the responses of a neural network to stimuli. This is because pathways can reveal the internal reactions and the working processes \citep{a04,a7}.

A pathway refers to a union of paths that go from an individual pixel of an input image to a feature map of a layer, including intermediate and output layers of a deep network. We define the pixels (regions) pointed by the large pathways as the parts of an object. And these pixels also pass through the important feature maps for classification (concept units). In this work, we introduce an algorithm to extract the pathways of individual pixels in a trained classification network. In this way, the intensity of a pixel spreading out to all feature maps through diffusion kernels can be visualized. The pathways allow us to test the causal components that are important for classification, indicating the location and contribution of an individual pixel from the input image to the feature maps it reaches. From this, we can deduce how much an image patch, called a part, contributes to a feature map and can establish a mapping between a part and an important unit (a learned feature map) for classification. And the part-based representation of all layers can provide us with detailed information about the internal responses of the data. Pathway-based neurotransmission also can contribute to our understanding of various aspects of biological neural networks (BNNs), such as the process of adversarial attacks \citep{a8}, the completion of occluded objects \citep{a9,a10,a11}, memory formation through synaptic connections \citep{a13,a14}, and perception of time intervals through object state evolution along a stereotypical trajectory or in working memory \citep{a15}. These perceptions may be influenced by the overlap or difference in the pathways over time.

We extract the pixel diffusion pathways from a VGG-16 convolutional neural network (CNN) \citep{a44} that has been trained on the MNIST \citep{a45} and CIFAR \citep{a46} datasets. We find that the few largest pathways of an individual pixel from an image tend to cross the feature maps in each layer that is important for classification. Additionally, there are only one or just a few pathways in the deep layers, while there are many pathways in the shallow layers. The ratio of the area of a part to the image constitutes the portion-hot codes in a given layer. The codes of the VGG-16, including 13 convolutional and 5 pooling layers, are concatenated to represent the internal processing of this image throughout the network, called the portion-hot representation. In the feature attribution experiments on intra- and inter-classes, we find that the main pathways of images belonging to the same category are more consistent in their trends than those of images from different categories, and the portion-hot centers of images from different categories are clearly distinguishable.
Further, we demonstrate the usefulness of our diffusion pathways in understanding of adversarial attacks, object completion, and movement perception. In experiments with adversarial samples from the CIFAR dataset generated by the fast gradient sign method (FGSM) \citep{a47}, we find that the portion-hot distances between the original and the target samples of different categories are significantly smaller than those between adversarial and the target samples of the same category. Similar experiments on cropped and rotated samples also reveal a comparable contrast for the portion-hot distances.

In summary, the main contributions of our work can be concluded as follows:

\begin{itemize}
	\item We present an algorithm to extract the pathways of individual pixels in a trained  classification network.
	\item This study shows that the pathways within classification networks can match a part of an object with locations in an image. 	
	\item Pathway-based representations explain why BNNs are competent for perception in adversarial attacks, object movement, and occlusion.
\end{itemize}

\section{Related Work}
Despite the recent advances in analyzing and interpreting deep neural networks, there are still several open questions and research gaps that need to be addressed. First, most of the existing methods focus on visualizing or explaining the behavior of individual units or layers \cite{1}, but they do not provide a comprehensive and systematic understanding of the network as a whole. Second, most of the existing methods are indirect, such as salience maps \cite{5,6} and surrogate models \cite{7, 8}. And some of them rely on external models or datasets to measure the agreement or similarity between network units and human concepts \cite{2,3,4}, but they do not account for the internal structure and function of the network itself. Some approaches of deep network interpretation follow the axiomatic attribution method \cite{a16, j5}, in which sensitivity and invariance are two fundamental axioms and the final gradient result does not depend on the intermediate details. These approaches are centered on the stimulus-response pattern and reflect an interest in discovering units or unit interactions that are important for a particular prediction by using a variety of effective methods \citep{a18,a20,j2,a30, j3, j4, j6}.

The pathway-based explanation considerers the overall internal mechanisms of the data processing and the location information of important feature maps. The data pathways \citep{a04,a7,a8} refer to a group of critical neurons along with their connections, which are identified by the pruning objective \citep{a7} or via the contribution neurons make to the response \citep{a04}. These data pathways need to be learned by building models. By contrast, the location in our pathway framework is different: it is explicit and fixed rather than implicit and learned. while our diffusion pathways are directly extracted without learning from a trained network for classification tasks.

%

\section{Extracting pathways in deep networks}
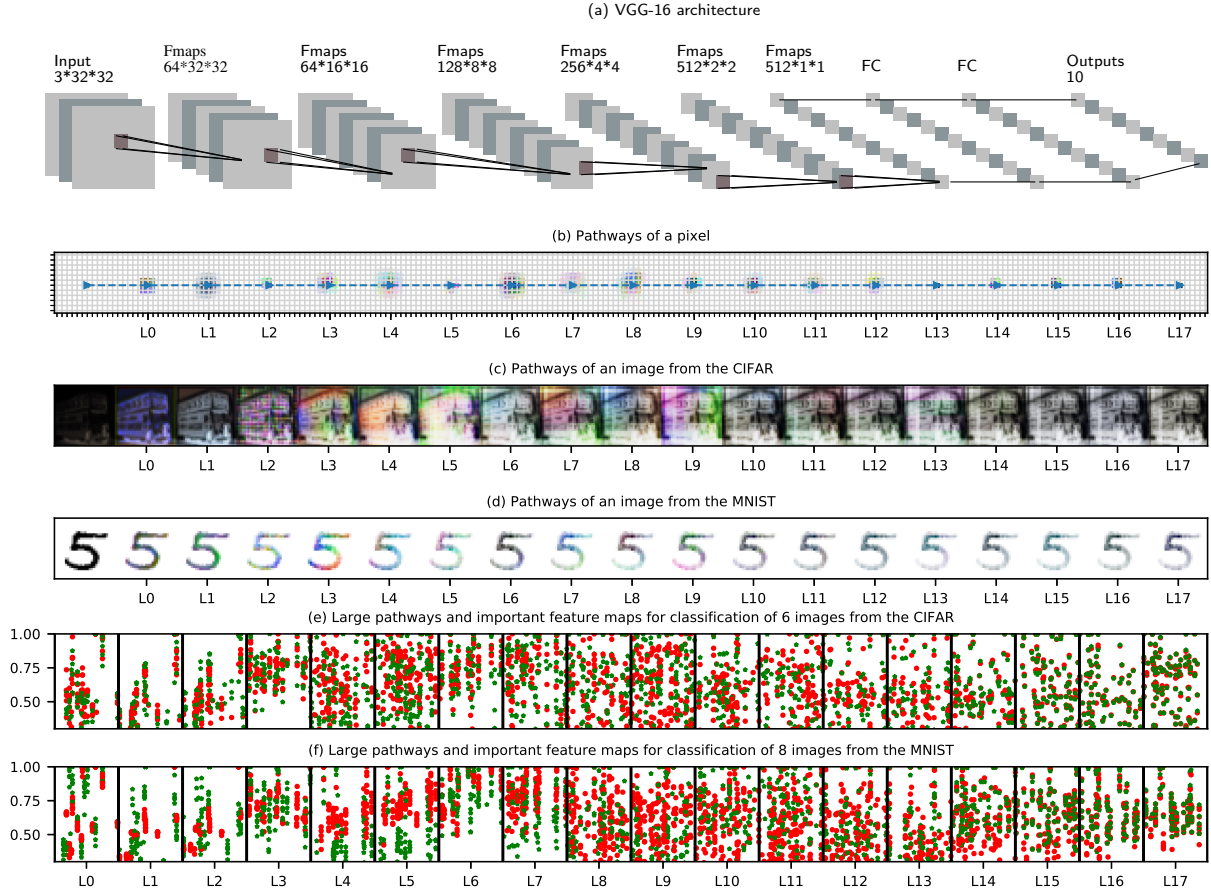
\begin{figure*}[htbp]
	\centering
	\vspace{-0.1em}	
	\setlength{\abovecaptionskip}{-0.2em}
	\setlength{\belowcaptionskip}{-0.1em}
	\begin{tikzpicture}[yscale=0.9,xscale=0.9]	
	\node [anchor=west,align=left,font=\fontsize{6}{6}\selectfont] at (7.8,23)
	{(a) VGG-16 architecture};
	\node [anchor=west,align=left,font=\fontsize{6}{6}\selectfont] at (0,22.2){  \\  Input \\ 3*32*32};	
	
	\begin{scope}[name prefix = l1-]
	\foreach \x/\y/\z in {0/20.6/first, 0.2/20.5/second,  0.4/20.4/first}
	\fill[\z]  (\x,\y) rectangle (\x+1.2,\y+1.2);
	\fill[third] (1,21) rectangle (1.2,21.2);
	\node [] (a1) at (0.9,21.2){};
	\node [] (a2) at (1,21.2){};
	\node [] (a3) at (0.9,21){};
	\node [] (a4) at (1,21){};
	\end{scope}
	
	\node [anchor=west,align=left,font=\fontsize{6}{6}\rmfamily\selectfont] at (1.6,22.3){ Fmaps\\64*32*32};
	\begin{scope}[name prefix = l2-]
	\foreach \x/\y/\z in {1.8/20.8/first, 2/20.7/second,  2.2/20.6/first, 2.4/20.5/second,  2.6/20.4/first}
	\fill[\z]  (\x,\y) rectangle (\x+1,\y+1);
	\fill[third] (3.2,20.8) rectangle (3.4,21);
	\node [] (b1) at (3.1,21){};
	\node [] (b2) at (3.3,21){};
	\node [] (b3) at (3.1,20.8){};
	\node [] (b4) at (3.3,20.8){};	
	\node [] (b0) at (3,20.8){};			
	\end{scope}		
	\draw [black] (l1-a1) -- (l2-b0);
	\draw [black] (l1-a2) -- (l2-b0);
	\draw [black] (l1-a3) -- (l2-b0);
	\draw [black] (l1-a4) -- (l2-b0);
	
	\node [anchor=west,align=left,font=\fontsize{6}{6}\selectfont] at (3.6,22.3){Fmaps\\64*16*16};
	\begin{scope}[name prefix = l3-]
	\foreach \x/\y/\z in {3.7/21/first, 3.9/20.9/second,  4.1/20.8/first, 4.3/20.7/second,  4.5/20.6/first, 4.7/20.5/second,  4.9/20.4/first}
	\fill[\z]  (\x,\y) rectangle (\x+0.8,\y+0.8);
	\fill[third] (5.2,20.8) rectangle (5.4,21);
	\node [] (c1) at (5.1,21){};
	\node [] (c2) at (5.3,21){};
	\node [] (c3) at (5.1,20.8){};
	\node [] (c4) at (5.3,20.8){};			
	\node [] (c0) at (5.2,20.6){};			
	\end{scope}
	
	\draw [black] (l2-b1) -- (l3-c0);
	\draw [black] (l2-b2) -- (l3-c0);
	\draw [black] (l2-b3) -- (l3-c0);
	\draw [black] (l2-b4) -- (l3-c0);
	
	\node [anchor=west,align=left,font=\fontsize{6}{6}\selectfont] at (5.6,22.3){Fmaps\\128*8*8};
	\begin{scope}[name prefix = l4-]
	\foreach \x/\y/\z in {5.8/21.2/first, 6/21.1/second,  6.2/21/first, 6.4/20.9/second,  6.6/20.8/first, 6.8/20.7/second,  7/20.6/first, 7.2/20.5/second,  7.4/20.4/first}
	\fill[\z]  (\x,\y) rectangle (\x+0.6,\y+0.6);	
	\fill[third] (7.8,20.6) rectangle (8,20.8);
	\node [] (d1) at (7.7,20.8){};
	\node [] (d2) at (7.9,20.8){};
	\node [] (d3) at (7.7,20.6){};
	\node [] (d4) at (7.9,20.6){};
	
	\node [] (d0) at (7.8,20.6){};
	
	\end{scope}
	\draw [black] (l3-c1) -- (l4-d0);
	\draw [black] (l3-c2) -- (l4-d0);
	\draw [black] (l3-c3) -- (l4-d0);
	\draw [black] (l3-c4) -- (l4-d0);
	
	\node [anchor=west,align=left,font=\fontsize{6}{6}\selectfont] at (7.4,22.3){Fmaps\\256*4*4};
	\begin{scope}[name prefix = l5-]
	\foreach \x/\y/\z in {7.6/21.4/first, 7.8/21.3/second,  8/21.2/first, 8.2/21.1/second,  8.4/21.0/first, 8.6/20.9/second,  8.8/20.8/first, 9/20.7/second, 9.2/20.6/first, 9.4/20.5/second,  9.6/20.4/first}
	\fill[\z]  (\x,\y) rectangle (\x+0.4,\y+0.4);
	
	\fill[third] (9.8,20.4) rectangle (10,20.6);
	
	\node [] (e1) at (9.7,20.6){};
	\node [] (e2) at (9.9,20.6){};
	\node [] (e3) at (9.7,20.4){};
	\node [] (e4) at (9.9,20.4){};
	\node [] (e0) at (9.8,20.7){};
	\end{scope}
	
	\draw [black] (l4-d1) -- (l5-e0);
	\draw [black] (l4-d2) -- (l5-e0);
	\draw [black] (l4-d3) -- (l5-e0);
	\draw [black] (l4-d4) -- (l5-e0);
	
	\node [anchor=west,align=left,font=\fontsize{6}{6}\selectfont] at (9.1,22.3){Fmaps\\512*2*2};	
	\begin{scope}[name prefix = l6-]
	\foreach \x/\y/\z in {9.3/21.5/first, 9.5/21.4/second, 9.7/21.3/first, 9.9/21.2/second,  10.1/21.1/first, 10.3/21/second,  10.5/20.9/first, 10.7/20.8/second, 10.9/20.7/first, 11.1/20.6/second,  11.3/20.5/first, 11.5/20.4/second}
	\fill[\z]  (\x,\y) rectangle (\x+0.3,\y+0.3);
	
	\fill[third] (11.6,20.4) rectangle (11.8,20.6);
	\node [] (f1) at (11.5,20.6){};
	\node [] (f2) at (11.7,20.6){};
	\node [] (f3) at (11.5,20.4){};
	\node [] (f4) at (11.7,20.4){};
	\node [] (f0) at (11.7,20.5){};
	\end{scope}
	
	\draw [black] (l5-e1) -- (l6-f0);
	\draw [black] (l5-e2) -- (l6-f0);
	\draw [black] (l5-e3) -- (l6-f0);
	\draw [black] (l5-e4) -- (l6-f0);
	
	\node [anchor=west,align=left,font=\fontsize{6}{6}\selectfont] at (10.4,22.3){Fmaps\\512*1*1};	
	\begin{scope}[name prefix = l7-]
	\foreach \x/\y/\z in {10.6/21.6/first, 10.8/21.5/second,  11/21.4/first, 11.2/21.3/second,  11.4/21.2/first, 11.6/21.1/second,  11.8/21.0/first, 12/20.9/second, 12.2/20.8/first, 12.4/20.7/second,  12.6/20.6/first, 12.8/20.5/second,  13/20.4/first}
	\fill[\z]  (\x,\y) rectangle (\x+0.2,\y+0.2);
	
	\node [] (g0) at (13.2,20.5){};
	
	\node [] (gfirst) at (10.6,21.7){};
	
	\node [] (glast) at (13.1,20.5){};
	
	\end{scope}
	
	\draw [black] (l6-f1) -- (l7-g0);
	\draw [black] (l6-f2) -- (l7-g0);
	\draw [black] (l6-f3) -- (l7-g0);
	\draw [black] (l6-f4) -- (l7-g0);

	\node [anchor=west,align=left,font=\fontsize{6}{6}\selectfont] at (11.8,22.2){FC};	
	\begin{scope}[name prefix = l8-]
	\foreach \x/\y/\z in {12/21.6/first, 12.2/21.5/second,  12.4/21.4/first, 12.6/21.3/second,  12.8/21.2/first, 13/21.1/second,  13.2/21.0/first, 13.4/20.9/second, 13.6/20.8/first, 13.8/20.7/second,  14/20.6/first, 14.2/20.5/second,  14.4/20.4/first}
	\fill[\z]  (\x,\y) rectangle (\x+0.2,\y+0.2);
	
	\node [] (hfirst1) at (12.2,21.7){};
	
	\node [] (hlast1) at (14.6,20.5){};
	
	\node [] (hfirst2) at (12.0,21.7){};
	
	\node [] (hlast2) at (14.4,20.5){};
	\end{scope}
	
	\draw [black] (l7-gfirst) -- (l8-hfirst1);
	\draw [black] (l7-glast) -- (l8-hlast1);
	%
	\node [anchor=west,align=left,font=\fontsize{6}{6}\selectfont] at (13.2,22.2){FC};	
	\begin{scope}[name prefix = l9-]
	\foreach \x/\y/\z in {13.4/21.6/first, 13.6/21.5/second,  13.8/21.4/first, 14/21.3/second,  14.2/21.2/first, 14.4/21.1/second,  14.6/21.0/first, 14.8/20.9/second, 15/20.8/first, 15.2/20.7/second,  15.4/20.6/first, 15.6/20.5/second,  15.8/20.4/first}
	\fill[\z]  (\x,\y) rectangle (\x+0.2,\y+0.2);
	
	\node [] (ifirst1) at (13.6,21.7){};
	\node [] (ilast1) at (16,20.5){};
	\node [] (ifirst2) at (13.4,21.7){};
	\node [] (ilast2) at (15.8,20.5){};
	\end{scope}
	\draw [black] (l8-hfirst2) -- (l9-ifirst1);
	\draw [black] (l8-hlast2) -- (l9-ilast1);
	
	\node [anchor=west,align=left,font=\fontsize{6}{6}\selectfont] at (14.8,22.2){\\Outputs \\10};
	\begin{scope}[name prefix = l10-]
	\foreach \x/\y/\z in {15/21.6/first, 15.2/21.5/second,  15.4/21.4/first, 15.6/21.3/second,  15.8/21.2/first, 16/21.1/second,  16.2/21/first, 16.4/20.9/second,16.6/20.8/first,16.8/20.7/second}
	\fill[\z]  (\x,\y) rectangle (\x+0.2,\y+0.2);
	
	\node [] (jfirst) at (15.2,21.7){};
	\node [] (jlast) at (17,20.8){};
	\end{scope}
	
	\draw [black] (l9-ifirst2) -- (l10-jfirst);
	\draw [black] (l9-ilast2) -- (l10-jlast);
	
	\end{tikzpicture}						
	
	
	\begin{minipage}{\linewidth}
		\centering	
		\includegraphics[width=0.97\textwidth]{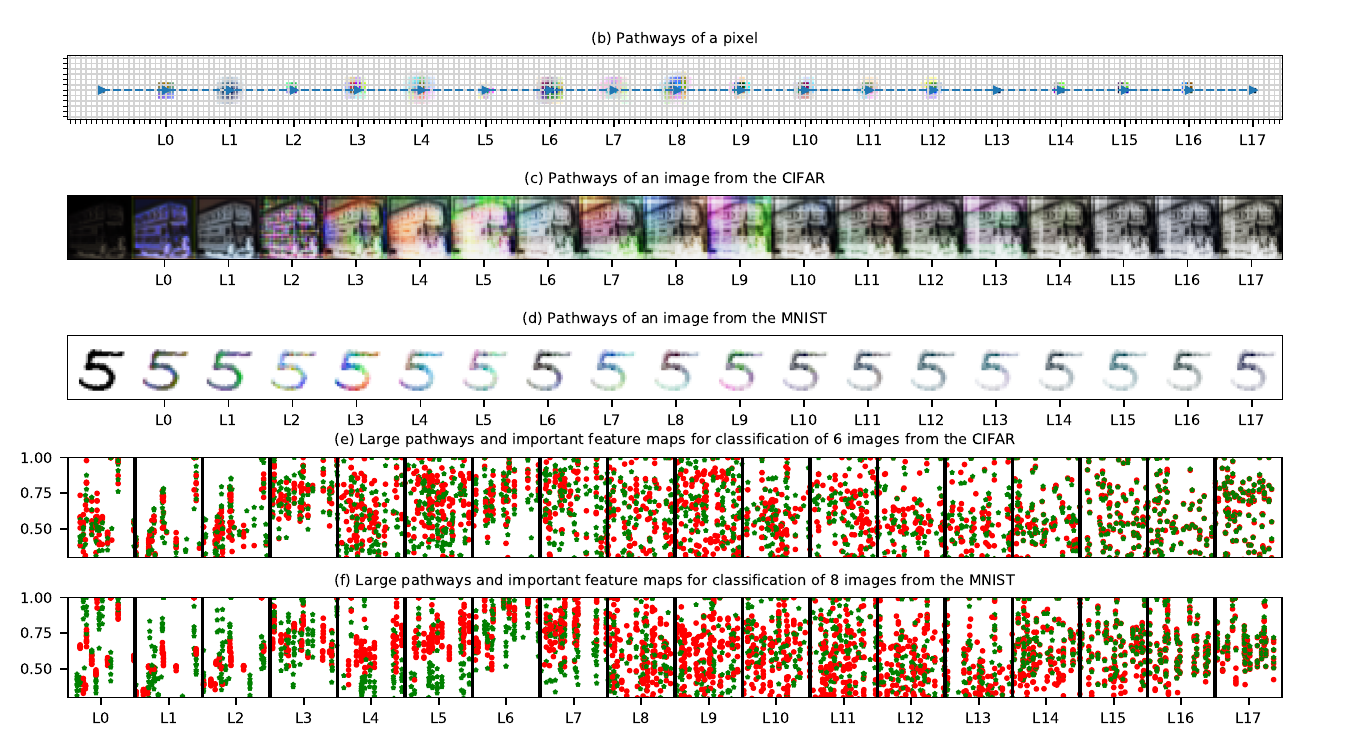}
	\end{minipage}
	\caption{The diffusion pathways of an individual pixel and of an image within the VGG-16 CNN, which was trained using the CIFAR and MNIST datasets, and the comparison between large pathways and important feature maps for classification in each layer. (a) The VGG-16 CNN consists of 13 convolutional layers (conv1\_1 through conv5\_3) and 5 max pooling layers (maxpl1 through maxpl5), followed by 3 fully connected layers. (b) Diffusion pathways of an individual pixel, with each showing the first three largest pathway cross-sections in the feature map in each of the 18 layers of conv1\_1 through maxpl5, from left to right, which are respectively marked by L0, L1, …, L17, and the same below. Diffusion pathways of an image randomly selected from the CIFAR dataset (c) and the MNIST dataset (d), with each showing the first three largest pathway cross-sections. The original image is in the first position. The distribution comparison between large pathway cross-sections (green) without channel mask constraints and important feature maps for classification (red) in each layer for a few images from the CIFAR dataset (e) and the MNIST dataset (f), and the Y-axis indicates the normalized intensity.}
	\label{fig1}	
\end{figure*}

We take a VGG-16 CNN (Fig. \ref{fig1}a) as an example to demonstrate how to extract pixel diffusion pathways across convolutional layers. The VGG-16 is trained to classify images into 10 categories using the CIFAR and the MNIST datasets. To analyze segmentation, the VGG-16 is also trained to classify the multilabel dataset M2NIST . 

\subsection{An algorithm for extracting diffusion pathways of individual pixels}
Diffusion kernels are well-trained convolutional filters for classification that are rotated 180 degrees around the channel axis. Let $ W(cout, cin, k, k) $ denote a filter for classification, and let $ RW(cout, cin, k, k) = rotate180(W) $ denote a diffusion kernel. The diffusion kernel does not violate the connectivity and data flow of the original classification network. The pathways of an individual pixel $ x_i $ start from position $ i $ of an image $ X(H, W, C), C = 1$ or $3 $ to all feature maps of given layers. The number of pathways corresponds to the number of feature maps, where the feature map can be viewed as a pathway cross-section. And the intensity of the pathways reflects the contribution of the pixel to this feature map.  

Pathway cross-sections of a pixel between layers are computed by the product of the diffusion kernel and the input.
\begin{equation}
\centering
Input: PS(cin,phin,pwin)  
\label{equ1}
\end{equation}

\begin{equation}
\begin{split}
Output: \sum_{cin} (Input \times (RW(cout,cin,k,k)+1))\\
 \rightarrow PS(cout,ph,pw)
\end{split}
\end{equation}
where the input is initialized as (C,1,1). $ph=phin+\lceil k/2 \rceil, pw=pwin+\lceil k/2 \rceil$.

The diffusion field of a pixel in a feature map is $ ph \times pw $. Each neuron of the diffusion field diffuses through the same kernel to a feature map in the next layer. 
The kernel plus $ 1 $ signifies that the cumulative pathway progressively intensifies from the input layer to subsequent layers, section by section.

The activation and pooling layers are transformed into masks to control the paths, turning them on or off. The activation mask $ ReluMask $ is correspondingly taken from the feature maps in the original activation layer, as is the pooling mask $ PoolMask $. 
\begin{equation}
ReluMask(PS(cout,ph,pw)) \rightarrow PS(cout,ph,pw)
\end{equation}

\begin{equation}
\begin{split}
PoolMask(PS(cout,ph,pw)) \rightarrow PS(cout,ph,pw),\\
ph=\lceil ph/2 \rceil, pw=\lceil pw/2 \rceil
\end{split}
\end{equation}

To further maintain the consistency of the pathways and the classification, we use the important feature maps for classification as a channel mask $ ChMask $ to constrain the pathways in each layer.
\begin{equation}
ChMask(PS(cout,ph,pw)) \rightarrow PS(cout,ph,pw)
\end{equation}

\begin{algorithm*}[htbp]
	
	\caption{An algorithm of diffusion pathways for a batch of images}
	\textbf{Input}: $PS=X(batch,H,W,C,ph,pw), ph=1, pw=1$\\
	\vspace{-\baselineskip} 
	\begin{algorithmic}[1] 
		\STATE Let $layer=0 $ 
		\WHILE{$layer<18$}
		\IF {$convolutional \quad layer$}
		\STATE 
		\begin{align*}		
		&\sum_{cin} (PS(batch,H,W,cin,ph,pw) \times (RW(cout,cin,k,k)+1))\\
		&\rightarrow PS(batch,H,W,cout,ph,pw), ph+=\lceil k/2 \rceil, pw+=\lceil k/2 \rceil;\\		
		&ReluMask(PS(batch,H,W,cout,ph,pw))\\ 
		&\rightarrow PS(batch,H,W,cout,ph,pw);\\	
		&ChMask(PS(batch,H,W, cout, ph, pw))\\ 
		&\rightarrow PS(batch,H,W,cout, ph, pw);
		\end{align*}                   	       
		\ELSIF {$pool \quad layer$}
		\STATE \begin{align*}	 
		&PoolMask(PS(batch,H,W,cout,ph,pw)) \\
		&\rightarrow PS(batch,H,W,cout,ph,pw), ph = \lceil ph/2 \rceil, pw = \lceil pw/2 \rceil
		\end{align*}
		\ENDIF
		\ENDWHILE
		\STATE \textbf{return} $PS(batch,H,W,cout,ph,pw)$
	\end{algorithmic}
	\textbf{Output}: $ \sum_{(ph,pw)}(PS(batch,H,W,cout,ph,pw)) 
	\rightarrow PS(batch,H,W,cout) $
	
	\textbf{Parts}: $ argtopk_{cout}(PS(batch,H,W,cout)) 
	\rightarrow PS(batch,H,W,C_{part}) $
	
	\textbf{Saliency map}: $ max_{cout}(PS(batch,H,W,cout)) 
	\rightarrow heatmap(batch,H,W) $
	
	\textbf{Portion-hot representation}: $ [batch,rep(part_{ij})] $
	\label{a1}
	
\end{algorithm*}

The pathways of all pixels of an image independently cascade through each feature map of each layer from input to output. The pathways of all pixels of an image to a layer are represented as $ PS(H,W,cout,ph,pw) $. We are interested in the overall effect of the pathways of the image on each feature map of a particular layer:
\begin{equation}
\sum_{(ph,pw)}(PS(H,W,cout,ph,pw)) \rightarrow PS(H,W,cout)
\end{equation}

The patch of pixels in an image pointed by the large pathways makes up the shape of a concept or an object part that matches the feature maps important for classification. However, these patches do not always conform to what is commonly agreed upon, except through learning guided by prior knowledge. Patches may intersect or overlap, i.e., one pixel may be shared by several patches. The patches can be obtained by aggregating several large pathways along the channel dimension:
\begin{equation}
argtopk_{cout}(PS(H,W,cout)) \rightarrow PS(H,W,C_{part})
\end{equation}

The saliency map, as an overall feature attribution, is obtained by retrieving the largest pathways along the channel dimension, assuming that there is no overlap between these patches.
\begin{equation}
max_{cout} (PS(H,W,cout)) \rightarrow heatmap(H,W)
\end{equation} 

The area ratio $ part_{ij} \in [0,1] $ of patch $ i \in ([0,63]/[0,127]\\/[0,255]/[0,511]) $ to an image displays its contribution to feature map $ i $ in layer $ j $, and the portion-hot representation is expressed by concatenating $ part_{ij} $ throughout the 5696 feature maps of the 18 layers from the first convolutional layer conv1\_1 to the final max pooling layer maxpl5.

The algorithm for extracting diffusion pathways of a batch of images is summarized as Algorithm \ref{a1}.

\renewcommand{\dblfloatpagefraction}{0.9}

\section{Experiments}
\subsection{Data and Reproducibility}
The CIFAR dataset \citep{a46} consists of 60000 RGB images with size 32 $\times$ 32 in 10 categories, and each category has 6000 images. The categories are airplane, automobile, bird, cat, deer, dog, frog, horse, ship, and truck. There are 50,000 training images and 10,000 test images. The MNIST dataset \citep{a45} of handwritten digits has a training set of 60,000 examples and a test set of 10,000 examples. The digits have been size-normalized and centered in the fixed-size 28$\times$28 images. The M2NIST dataset \citep{a49} was generated by selecting up to 3 random 28$\times$28 grayscale images from the MNIST dataset and copying them into a single 64$\times$84 image. The digits were pasted so that they did not overlap, and no transformations were applied to the original images, ensuring that digits in the M2NIST dataset maintain the same orientation as they have in the MNIST dataset. The dataset has 5000 multidigit images with up to 3 digits.

The VGG-16, which is publicly available, loads the parameters of all layers directly from the PyTorch framework. The parameters are  pretrained using the images in the ImageNet dataset. The 13 convolutional layers of the VGG-16 are divided into 5 groups, each followed by a max pooling layer. In this paper, the VGG-16 is trained to classify images into 10 object categories using the CIFAR dataset and into 10 digital categories using the MNIST dataset, achieving classification accuracies of 83\% and 98\%, respectively, with the validation subsets of the datasets. It is also trained to classify the multilabel images using the M2NIST dataset, and the precision and the recall are 56\% and 58\%, respectively. It processes input images at 32$\times$32 resolution for the CIFAR dataset, 28$\times$28 for the MNIST, and 48$\times$48 for the M2NIST. The feature maps are pooled to halve the resolution at each successive pool layer so that the last pooling layer and the convolutional layer conv5\_3 contain 512 feature maps at a minimum 1$\times $1 resolution.We implemented these experiments with PyTorch on Dell Precision 7760 with 128GB RAM and 16GB GPU. 

The pathways extracted by Algorithm \ref{a1}, the adversarial samples generated by the FGSM, and the portion-hot representation data sets are pre-calculated and stored in folders before analysis.


\renewcommand{\dblfloatpagefraction}{0.9}
\begin{figure*}[htbp]
	\centering
	\includegraphics[width=0.97\textwidth]{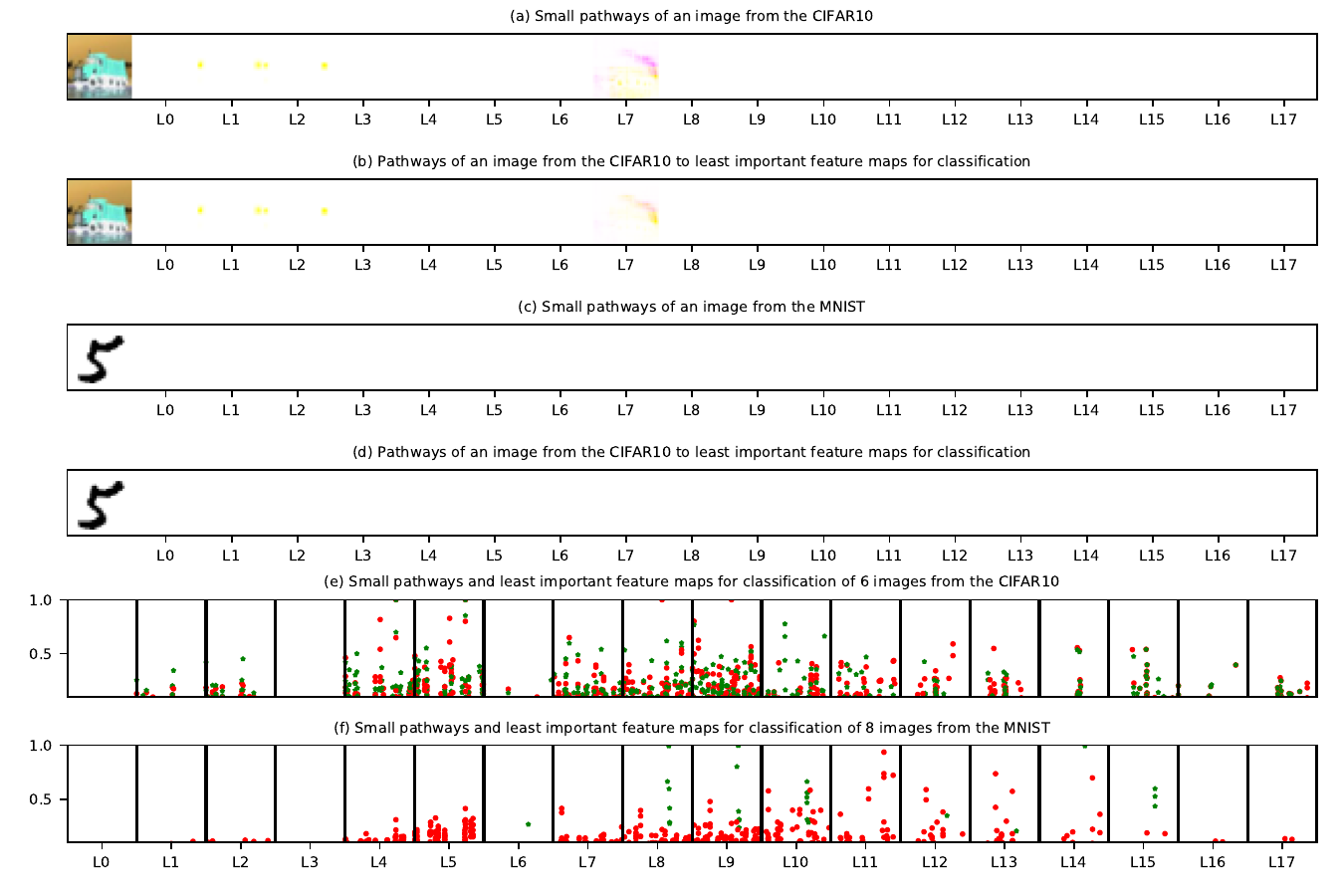}
	\caption{The diffusion pathways within the VGG-16 CNN, which was trained using the CIFAR and MNIST datasets, and the comparison between small pathways and least important feature maps for classification in each layer. Pathways of an image randomly selected from the CIFAR dataset (a) and the MNIST dataset (c): each showing the three smallest pathways in each of the 18 layers of 13 convolutional layers and 5 pooling layers. Diffusion pathways of an image randomly selected from the CIFAR dataset (b) and the MNIST dataset (d), with each showing the pathways at the three least important feature map for classification. The distribution comparison between small pathways without channel mask constraints and least important feature maps for classification in each layer for images from the CIFAR dataset (e) and the MNIST dataset (f). }
	\label{fig2}
\end{figure*}

\subsection{Displaying pathways}
Let's examine the process of extracting the diffusion pathways of an individual pixel and an image through the 13 convolutional layers (conv1\_1 to conv5\_3) and 5 max pooling layers (maxpl1 to maxpl5) of the VGG-16 model, as shown in Fig. \ref{fig1}a. The three largest pathway cross-sections (denoted by different colors) of one pixel in the 18 layers are shown from left to right in Fig. \ref{fig1}b, which alternately expand and contract from input to output due to the impact of activation and pooling. The size of its diffusion field in the feature map in the final pooling layer maybe 1 for very deep networks and small images.

It is known that the receptive field of a feature map is the overall image space, and the pathways from an image to the important feature maps also carry the overall spatial information. The first three largest pathway cross-sections of an image are shown in Fig. \ref{fig1}c and \ref{fig1}d which are randomly selected from the CIFAR dataset (Fig. \ref{fig1}c) and the MNIST dataset (Fig. \ref{fig1}d). These pathway cross-sections exhibit similar shapes in all layers because each pixel spreads through the same diffusion kernel, and its pathways are less disturbed by activation and pooling.

Because the pathways are optionally masked with the important feature maps for classification (channel masks), the intensity of the pathway cross-section of an image is not exactly equal to the intensity of the feature map for classification that it crosses. But there is a consistent distribution between them because the diffusion kernel is just the rotation of the convolution filter trained for classification. For images randomly selected from the CIFAR dataset (Fig. \ref{fig1}e) and the MNIST dataset (Fig. \ref{fig1}f), it can be seen that the green '*' denoting the first 10 largest pathway cross-sections without channel mask constraints almost cover the red 'o' denoting the first 10 largest feature maps for classification in each layer. 

As shown in Fig. \ref{fig2}, the pathways of an image do not provide overall spatial information to small pathways and to the least important feature maps for classification. However, the intensity distributions of the pathway cross-sections without channel mask constraints and the feature maps for classification in each layer are mostly consistent.

\begin{figure*}
	\centering
	
	\setlength{\abovecaptionskip}{-0.8em}
	\setlength{\belowcaptionskip}{-0.1em}
	\includegraphics[width=1\textwidth]{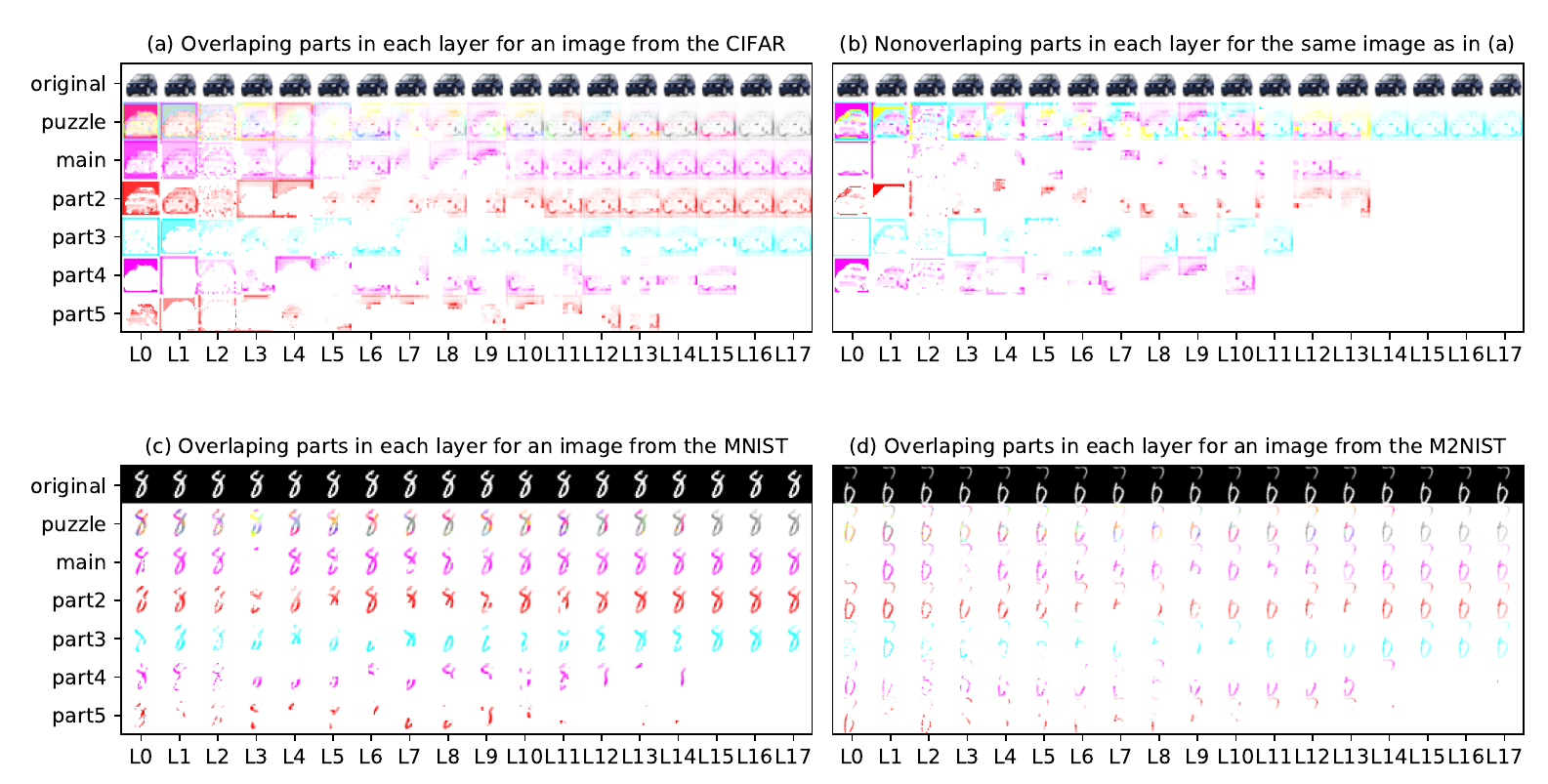}
	\caption{Part pathways and main pathways in each layer. The original image is in the first row, the puzzle made up of parts is in the second row, and the parts are in the third (the largest, i.e., the main part) to the seventh row, in descending order by area. The horizontal axis indicates the 18 layers, from left to right. An image randomly selected from the CIFAR dataset (a), from the MNIST dataset (c), and from the M2NIST dataset (d) is segmented into five parts from top to bottom by the large pathways in each layer. (b) Nonoverlapping parts for the same image as in (a).}
	\label{fig3}
	
\end{figure*}

\subsection{Parts and main pathways}

The object detectors in a classification network are used to discover which regions of an image activate the concept units\citep{a01}. In Fig. \ref{fig3}, we segment the image selected from the CIFAR dataset into five parts and depict the three largest pathways of each pixel in the 18 layers from left to right.
We can see that these five parts contain some overlapping content in Fig. \ref{fig3}a which indicates that one pixel may be shared by different parts. A part for each layer matches an important feature map (concept) of that layer. The segmented parts in each layer in turn create a puzzle from the image  (row 2). The number of segmented parts decreases from conv1\_1 to maxpl5. This is consistent with the unit-based analysis that information converges from being scattered to becoming concentrated from shallow layers to deep layers. While each feature map in every layer has a receptive field that encompasses the entire image, a shallow feature map is more likely to represent a specific part, whereas a deep feature map is more likely to capture a combination of multiple parts, forming an object. In the case of an image depicting a car, layer 2 typically consists of at least five parts, such as the car body (row 3) and the background (row 5). The last pooling layer typically contains only three parts. In Fig. \ref{fig2}b, non-overlapping parts depicts that the number of parts is reduced and the shape fullness of each part in each layer appears weaker. This phenomenon illustrates that the shape of the part generated by the pathways is not unique. 

In addition, we conduct the experiments on the  MNIST dataset (Fig. \ref{fig3}c) and M2NIST dataset \citep{a49} (Fig. \ref{fig3}d). The images in M2NIST dataset contain three separate digits. The experimental results show that, in the process of image cognition, the neural network learns the image as a whole, rather than segmented and then learned according to the individual in the image.


\renewcommand{\dblfloatpagefraction}{0.9}
\begin{figure*}[tbhp]
	\centering
	\vspace{0em}	
	\setlength{\belowdisplayskip}{0.em}
	\setlength{\abovedisplayskip}{0.em}
	\setlength{\abovecaptionskip}{-0.4em}
	\setlength{\belowcaptionskip}{-0.1em}
	\includegraphics[width=0.97\linewidth]{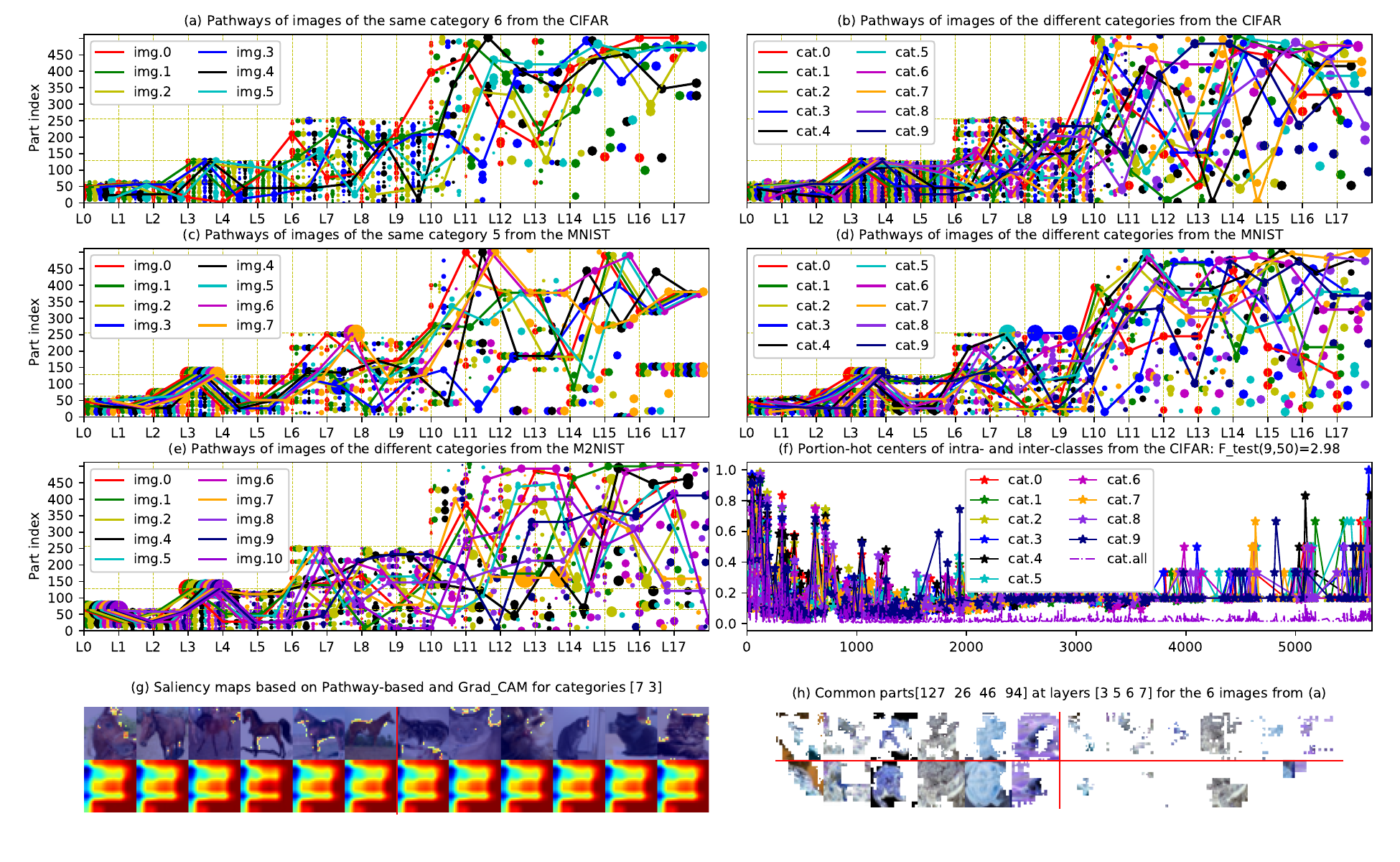}
	\caption{The main pathways and the portion-hot representations along the 18 layers for intra- and intercategories under accurate classification. (a) The scatter of parts (spots) and the main pathways (plots) of 6 images from the same category '6' randomly selected from the CIFAR dataset. The Y-axis denotes the part index (feature map index), and the X-axis denotes the 18 layers. The spot size indicates the area ratio of parts to the whole image. (b) The scatter of parts (spots) and the main pathways (plots) of 10 images from 10 categories randomly selected from CIFAR dataset. (c) The scatter of parts (spots) and the main pathways (plots) of 8 images from the same category '5' of the MNIST dataset. (d) The scatter of parts (spots) and the main pathways (plots) of 10 images from 10 categories of the MNIST dataset. (e) The scatter of parts (spots) and the main pathways (plots) of 10 multilabel images from the M2NIST dataset. (f) The centers of the portion-hot representations (5696 dimensions) of each category randomly selected from the CIFAR dataset and their average center (‘cat.all’). (g) The saliency maps from the pathway-based method (upper) and the Grad-CAM method (lower). (h) The common parts within the 6 images same as in (a) corresponding to the pathway sections in feature maps 127 of layer 3 (top left), 26 of layer 5 (top right), 46 of layer 6 (lower left), and 94 of layer 7 (lower right).}
	\label{fig4}
\end{figure*}

\subsection{Feature attribution by main pathways and portion-hot representations}

We understand how networks work internally by visualizing the comparison of the main pathways and the centers of the portion-hot representation in intra- and inter-categories.

Fig. \ref{fig4}a shows the scatter of the parts in each layer of 6 images randomly selected from category ‘6’ of the CIFAR dataset. Each image is assigned a different color. 
The plot connecting the largest spot of an image in each layer represents the main pathway of an image. 
The number of small spots in the shallow layers is relatively high, while the number of large spots in the deep layers is also high. This indicates that all pixels of an image tend to be concentrated in a few feature maps or parts of deeper layers, consistent with the conclusions obtained from part segmentation. Looking vertically, the main pathways of images of the same category coincide at shallow layers, such as layers 3 and 8. In contrast, the main pathways of images of different categories lack this sort of coincidence (Fig. \ref{fig3}b), and the centers of the portion-hot representations (5696 dimensions) of images with different categories are significantly different (Fig. \ref{fig3}f). The representation center of each category and the center of all categories (dark violet) are calculated for 60 images that are randomly selected, i.e., 6 images from each category of the CIFAR dataset. The center of all categories is significantly smaller at the deep dimensions, which again indicates that the center vector of each category is not concentrated in specific dimensions but scattering across different dimensions. In the analysis of variance (ANOVA) of these 60 samples, $ F_{test}=2.98 > F_{\alpha=0.05}(9,50)=2.08 $, and the difference in the portion-hot representations between categories is significant at the confidence level of 95\%. Similar results were presented on the MNIST and M2NIST datasets (Fig \ref{fig4}d, \ref{fig4}e). This demonstrates the generality of the pathways approach to understanding images.


The saliency map specifies the effect of key regions within an image on the prediction of this image. We use gradient-weighted class activation mapping (Grad-CAM) \citep{a43} as a baseline to evaluate the saliency map generated by the pathway-based method (Fig. \ref{fig3}g). The upper half of Fig. \ref{fig3}g shows the saliency maps produced by the largest pathways, and the lower half shows those produced by Grad-CAM. Since Grad-CAM is not capable of calculating saliency maps of small images, the saliency maps are produced based on the gradients on CONV3\_3. For 12 images of categories ‘7’ and ‘3’ from the CIFAR dataset, Grad-CAM produces the same saliency map, which is obviously wrong. On the contrary, the pathway-based saliency maps give accurate regions or locations. In different images of the same category, the common part in each layer may have different forms. Fig. \ref{fig4}h shows the shapes of the common parts within 6 images in 4 layers. For example, in the top left of Fig. \ref{fig4}h, the feature map ‘127’ of the maxpl1 layer (layer 3) corresponds to backgrounds in different orientations.

\renewcommand{\dblfloatpagefraction}{0.9}
\begin{figure*}[htbp]
	\centering
	\vspace{0em}	
	\setlength{\belowdisplayskip}{0.em}
	\setlength{\abovedisplayskip}{0.em}
	\setlength{\abovecaptionskip}{-0.2em}
	\setlength{\belowcaptionskip}{-0.1em}
	\begin{tikzpicture}[scale=0.9]
	\node [anchor=west,align=left,font=\fontsize{6}{6}\selectfont] 
	{(a) The three largest parts of three images};
	\end{tikzpicture}
	\vskip -5pt		 
	\begin{minipage}{\linewidth}	
		\includegraphics[width=0.97\linewidth]{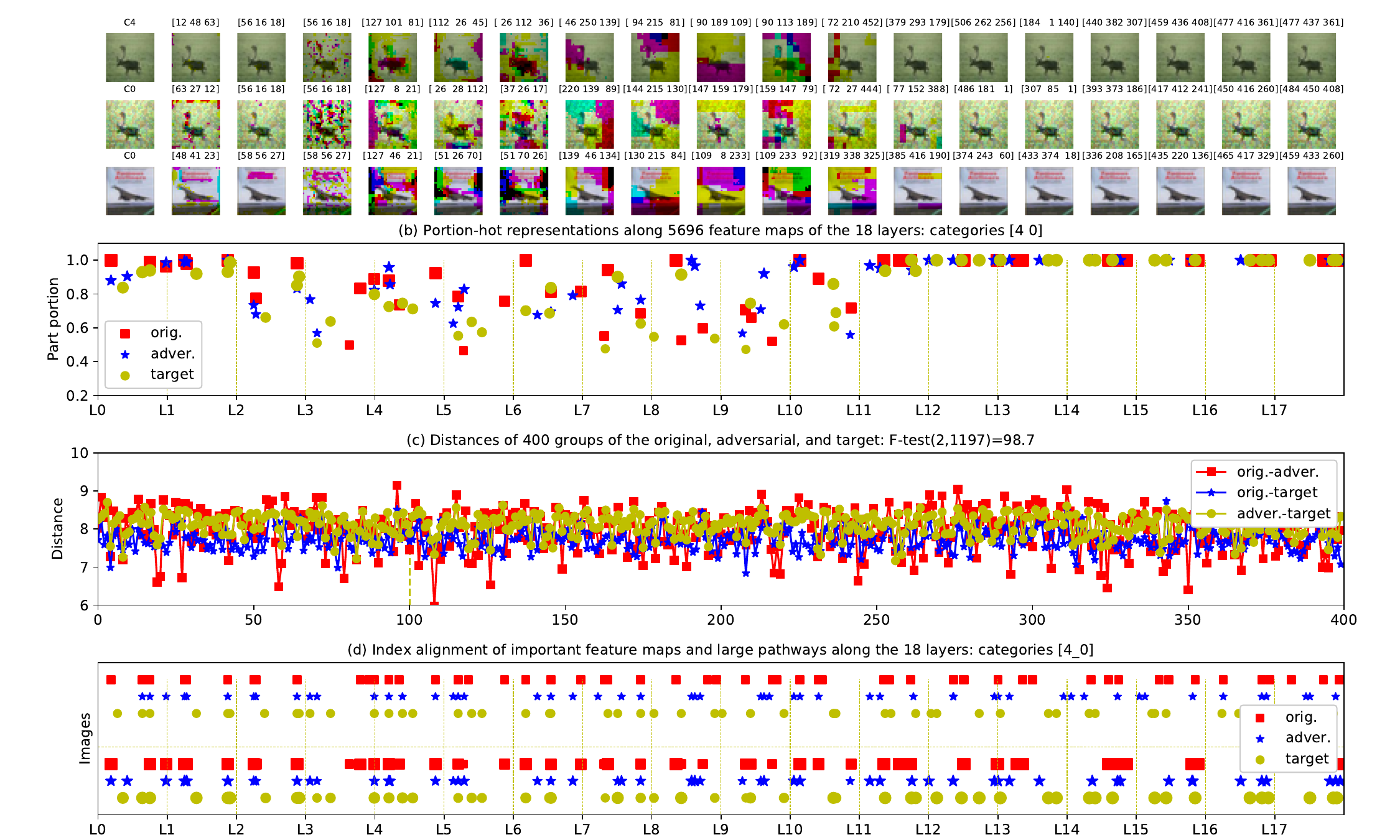}
	\end{minipage}
	\caption{Understanding the adversarial phenomenon through the part pathways and the portion-hot representations. (a) The three largest parts at each of the 18 layers of a group of three images, from top to bottom: an original image, an adversarial version of the original image, and a target image. Different colors mark the three parts at each layer. The three initial images are in the first column. The part indexes are at the top of each subplot. (b) The portion-hot representations of the group of three images: the original (red), the adversarial (blue), and the target (yellow). The X-axis indicates the 5696 feature maps of the 18 layers, and the Y-axis indicates the part area ratio to image. The marker size also denotes the area ratio of a part. (c) The distances of the portion-hot representation of 400 groups randomly selected from the CIFAR dataset and generated by the FGSM: the distances between the original and the adversarial (red), between the original and the target (blue), and between the adversarial and the target (yellow). The X-axis indicates the group number, and the Y-axis indicates the distance of the portion-hot representation. (d) The index alignment for the three images of the feature maps that are important for classification at the upper end and of the large part pathways at the lower end along the 18 layers, and the Y-axis indicates the three images. Note: To highlight the distribution of each layer, the layers are set to be equally spaced along the X-axis.}
	\label{fig5}
\end{figure*}
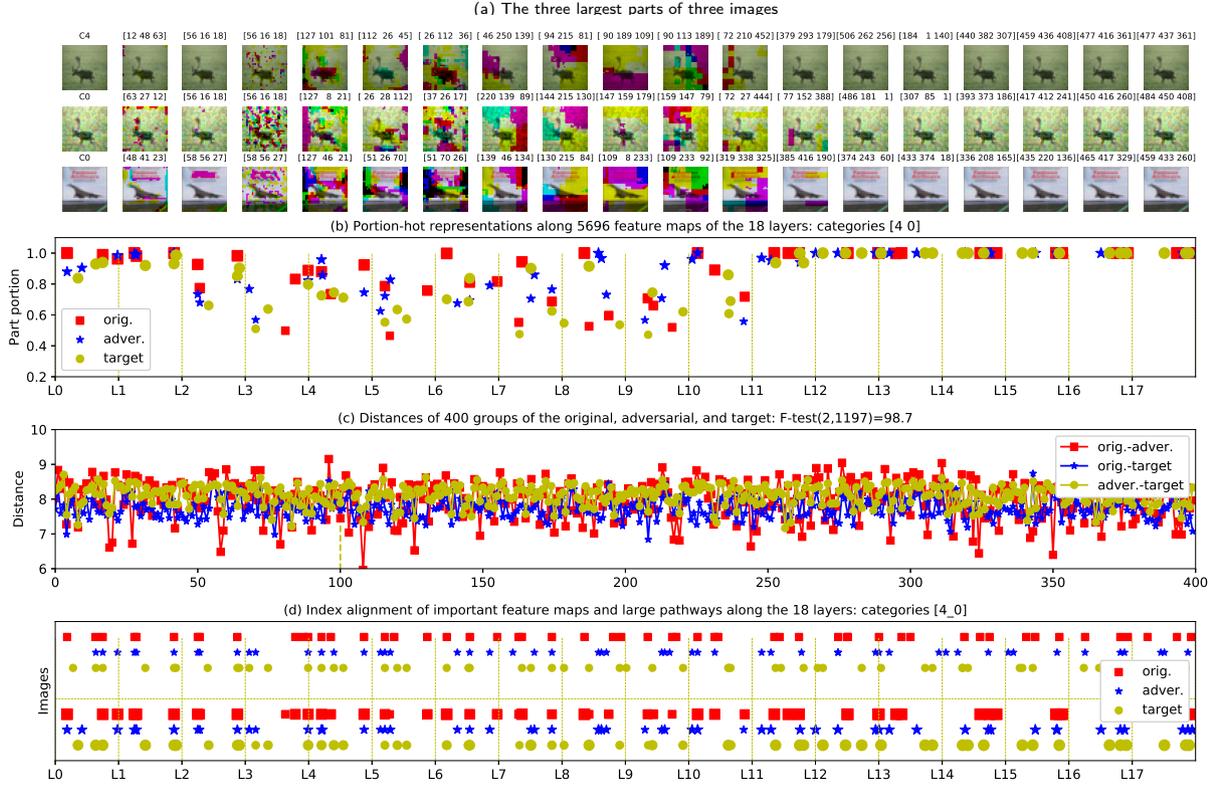

\section{Application: Adversarial attacks and perception of occlusion and movement}
\subsection{Adversarial attacks}
Figuring out how to analyze the adversarial attack phenomenon is an active area of research. Adversarial samples that will be misclassified by the classifier are generated by adding noise interference to an original sample. Although the small changes caused by each pixel do not produce visual differences, they can cause a sharp change in the L2 norm of the image, which can then cause the network to misclassify the image. We examine the pathways of a group of three images: an original image randomly selected from the CIFAR dataset, an adversarial version of that original image that is generated with the FGSM, and a target image belonging to the category of the adversarial version. In Fig. \ref{fig5}, a correctly classified ‘deer’ image (category 4) is attacked by the FGSM to be redirected toward the target ‘airplane’ (category 0). Fig. \ref{fig5}a shows the part pathways of the three images in the 18 layers. The part pathways of the adversarial image gradually approach the target and move away from the original. In the deep layers, although the part shapes look the same, their part indexes are different, which is depicted by the superscript of Fig. \ref{fig5}a. Their portion-hot representations along the 18 layers (5696 dimensions) are shown in Fig. \ref{fig5}b. The blue spots (adversarial), the red spots (original), and the yellow spots (target) cross in almost all layers. Where 400 groups of images are randomly selected from the CIFAR dataset and generated with the FGSM, the L2 distances of the portion-hot representation between the original and the target images (between different categories) are obviously smaller than the distances between the adversarial and the target images (between same category) (Fig. \ref{fig5}c).  In other words, there is a significant difference between the adversarial sample and the target sample although they are predicted by the classifier to be the same category. In the ANOVA, $ F_{test}=98.7 > F_{\alpha=0.05}(2,1197)=3.0 $, and the difference in the portion-hot representations among the three types of images is significant at a confidence level of 95\%.

The upper part of Fig. \ref{fig5}d shows the index alignment of the important feature maps for the classification of the three images. The lower part shows the index alignment of the large part pathways. In the upper and lower parts, the red squares and the yellow circles are almost aligned vertically, which once again proves that the large pathways and the important feature maps for classification are consistent with each other, even for adversarial attacks. By comparing the blue pentagrams in the upper and lower parts, we can find that the adversarial image gradually moves away from the original converging towards the target.

\renewcommand{\dblfloatpagefraction}{0.9}
\begin{figure*}[htbp]
	\centering
	\vspace{-0.1em}	
	\setlength{\abovecaptionskip}{-0.2em}
	\setlength{\belowcaptionskip}{-0.1em}
	\begin{tikzpicture}[scale=0.9]
	\node [anchor=west,align=left,font=\fontsize{6}{6}\selectfont] 
	{(a) The three largest parts of four images};
	\end{tikzpicture}
	\vskip -5pt		 
	\begin{minipage}{\linewidth}	
		\includegraphics[width=\linewidth]{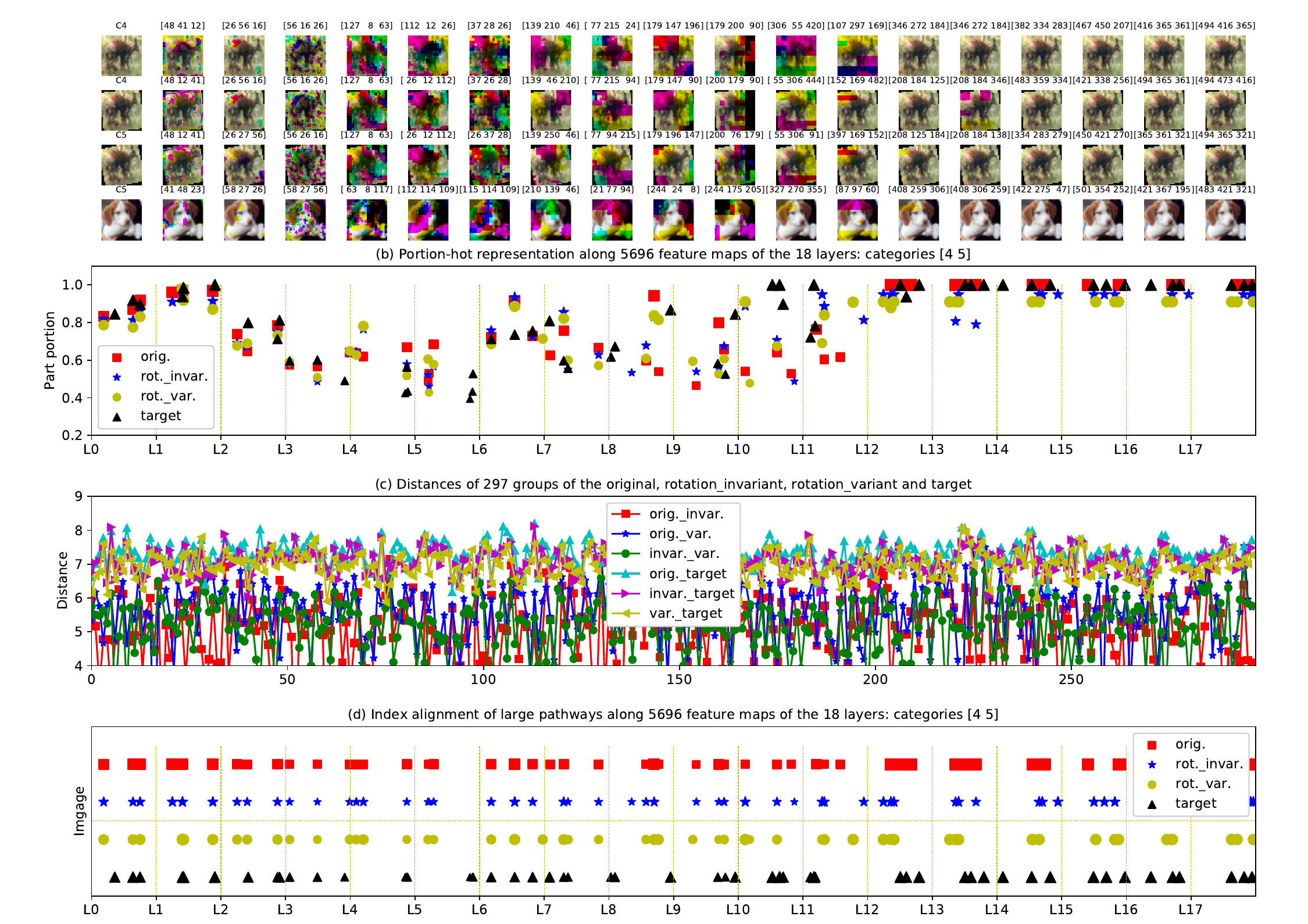}
	\end{minipage}
	\caption{Understanding an object movement by the part pathways and the portion-hot representations. (a) The three largest parts in each of the 18 layers of a group of four images, including, from top to bottom, an original image, its rotation-invariant version, its rotation-variant version, and the target version. Different colors mark the three parts in each layer. The four initial input images are in the first column. Part indexes are at the top of each subplot. (b) The portion-hot representations of the group of four images: the original (red), the rotation-invariant (blue), the rotation-variant (yellow), and the target (black). The X-axis indicates the 5696 feature map indexes of the 18 layers, and the Y-axis indicates the part portion. (c) The distances of the portion-hot representation of 297 groups: the distances between the original and its rotation-invariant image (red), the original and its rotation-variant image (blue), the invariant and the variant (green), the original and the target (cyan), the invariant and the target (magenta), the invariant and the target (yellow). The X-axis indicates the group number, and the Y-axis indicates the distance of the portion-hot representation. (d) The index alignment of the large parts along the 18 layers for the four images. The X-axis indicates the 5696 feature map indexes of the 18 layers, and the Y-axis indicates the four images. The marker size denotes the area ratio of a part in (b) and (d).}
	\label{fig6}
\end{figure*}
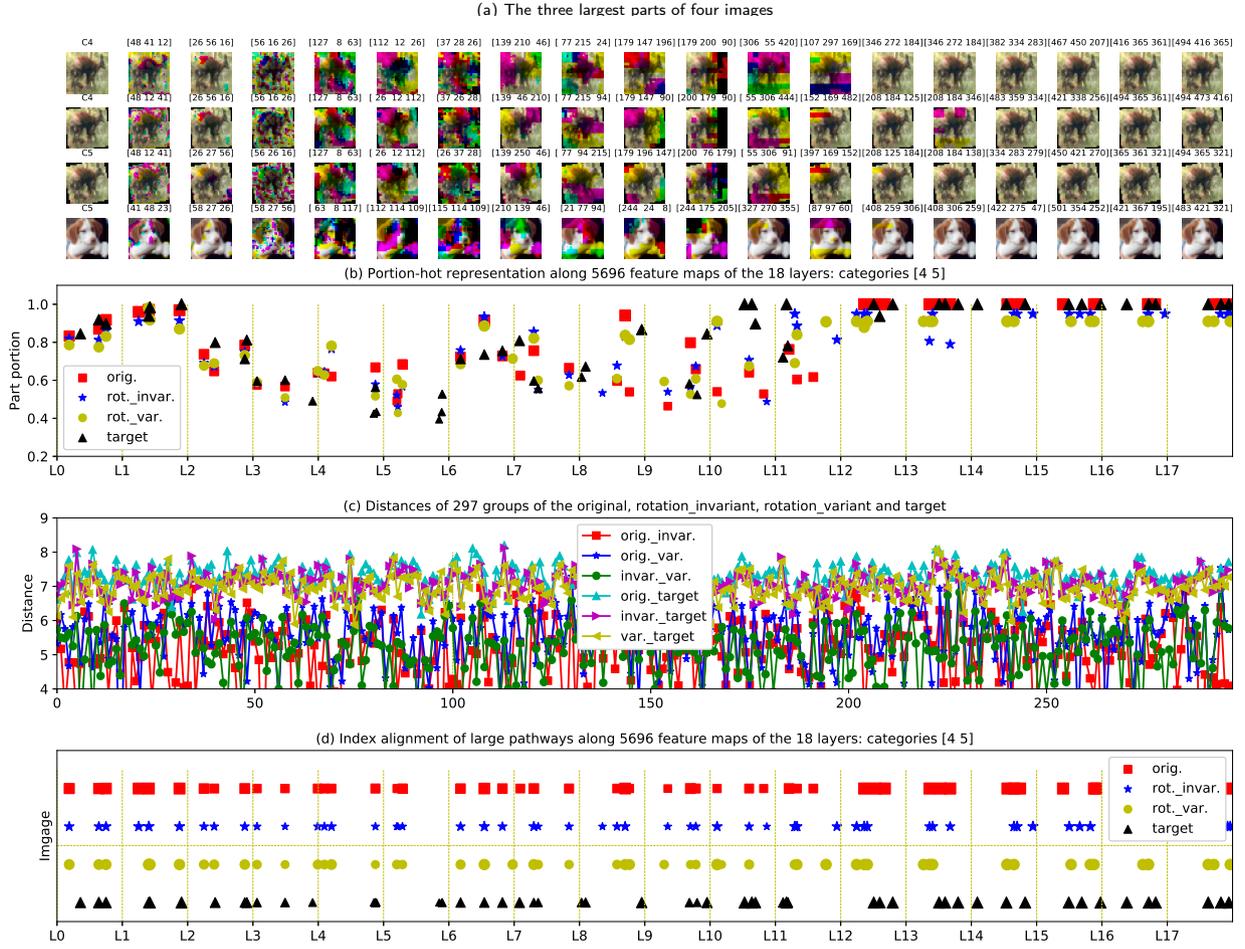

\subsection{Perception of object movement and occlusion}
BNNs perceive moving and occluded objects through object state evolution along a stereotypical trajectory or in working memory. And the pathway fit of an image and its transformed version can help us gain a preliminary understanding of these two perception mechanisms in BNNs.

The part pathways of the transformed image help us to know the network's ability to sense the movement of objects. Taking the rotation as an example, we examine the part pathways of a group of four images: an original image, its rotation-invariant version whose category label does not change, its rotation-variant version whose category label has changed, and the target version belonging to the category of its rotation variant. As shown in Fig. \ref{fig6}, the six distances of the portion-hot representation between four images per group are clearly divided into two types, and the three distances between the original sample and its two rotated versions are shorter than the three distances between the original sample and the target of different categories. In other words, there are significant differences between other categories and the original category and its variations.

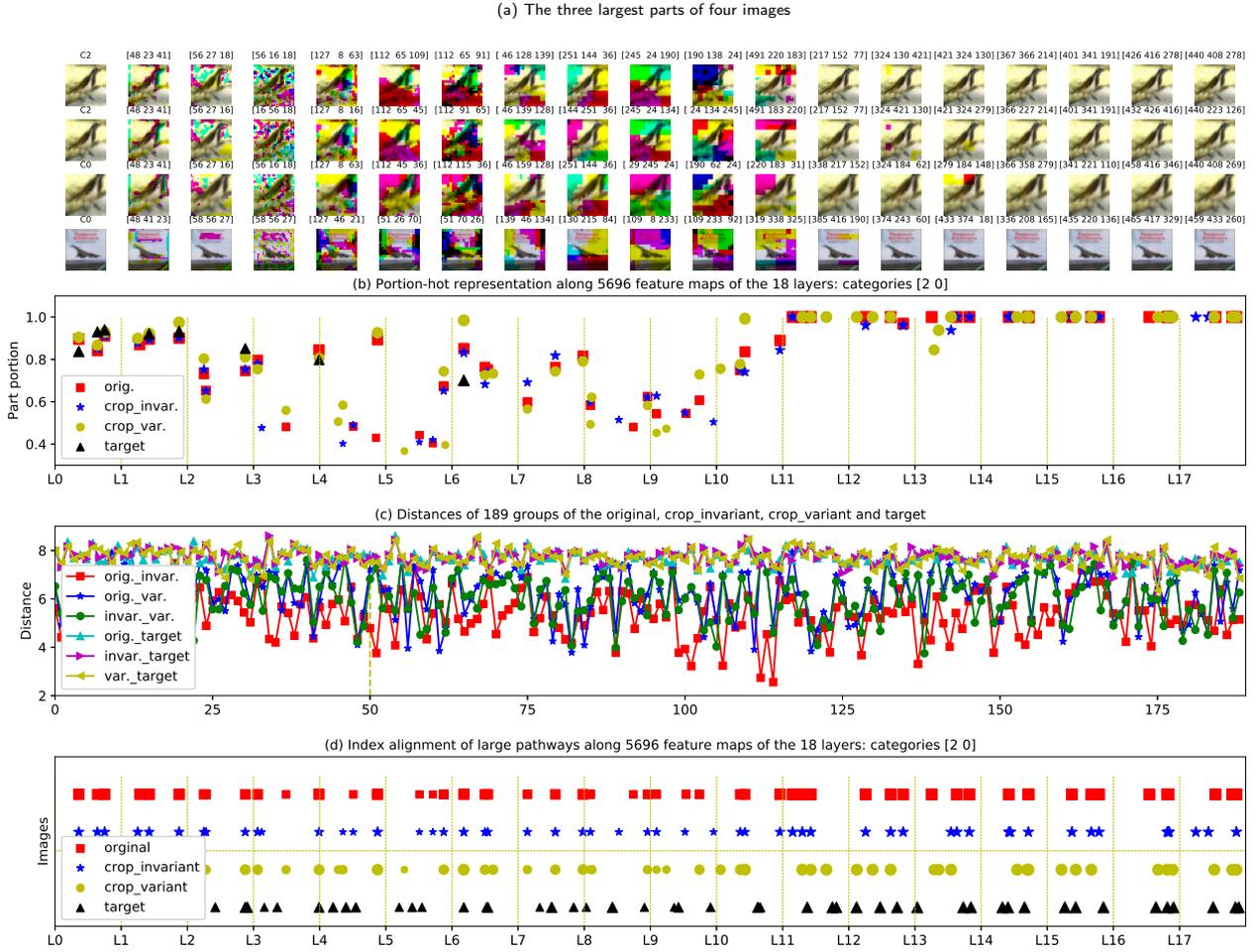
\begin{figure*}[htbp]
	\centering
	\vspace{-0.1em}	
	\setlength{\abovecaptionskip}{-0.2em}
	\setlength{\belowcaptionskip}{-0.1em}
	\begin{tikzpicture}[scale=0.9]
	\node [anchor=west,align=left,font=\fontsize{6}{6}\selectfont] at ([xshift=0.35\paperwidth ,yshift=-0.1\paperheight]current page.north west)
	{(a) The three largest parts of four images};
	\end{tikzpicture}		 
	\begin{minipage}{\linewidth}	
		\includegraphics[width=\linewidth]{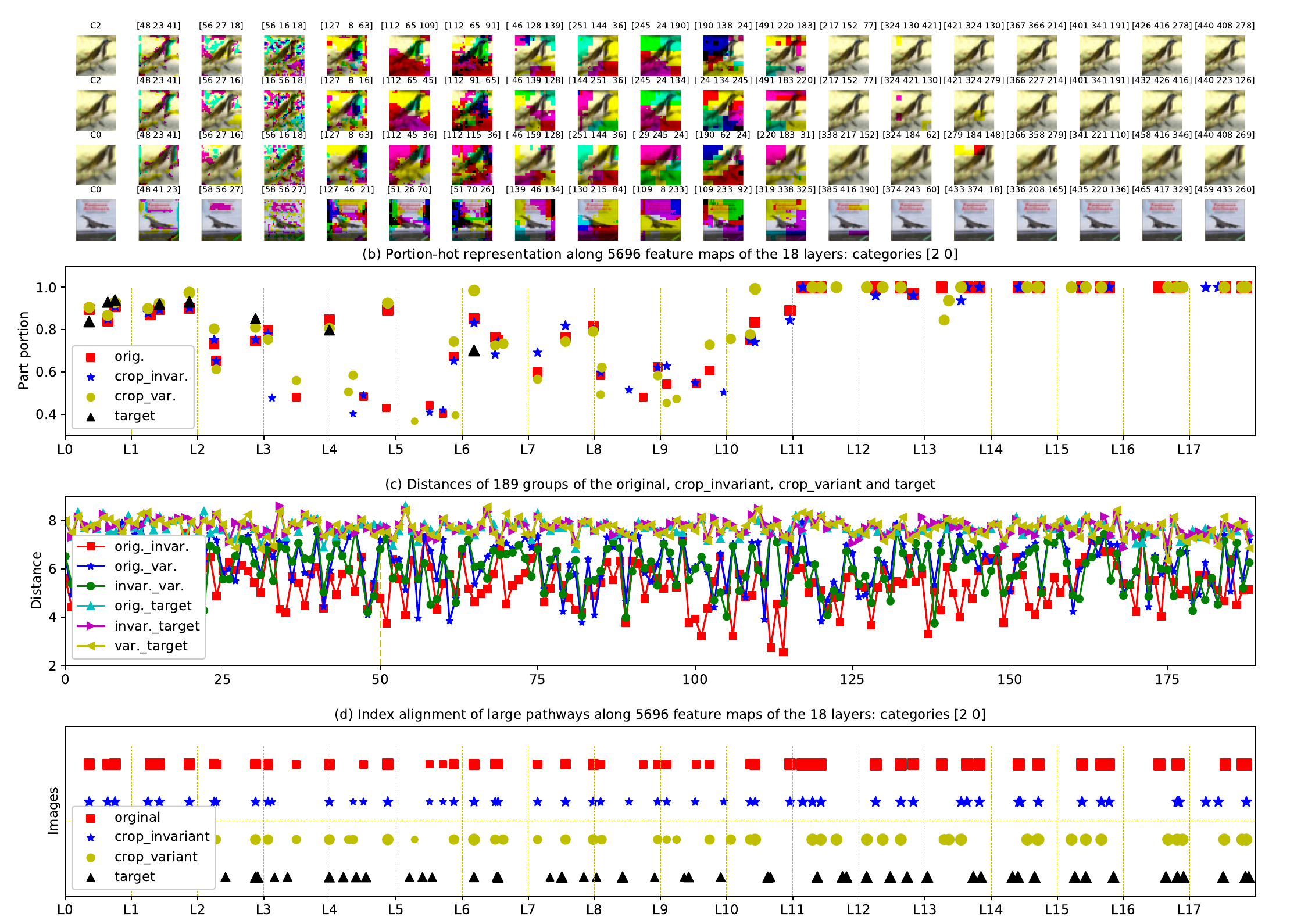}
	\end{minipage}
	\caption{Understanding the completion recognition by the part pathways and the portion-hot representations. (a) The three largest parts in each of the 18 layers of a group of four images, including, from top to bottom, an original image, its crop-invariant version, the crop-variant version, and the target version. Different colors mark the three parts in each layer. The four initial input images are in the first column. Part indexes are at the top of each subplot. (b) The portion-hot representations of the group of four images: the original (red), the crop-invariant (blue), the crop-variant (yellow), and the target (black). The X-axis indicates the 5696 feature map indexes of the 18 layers, and the Y-axis indicates the part portion. (c) The distances of the portion-hot representation of 189 groups: the distances between the original and its crop-invariant image (red), the original and its crop-variant image (blue), the invariant and the variant (green), the original and the target (cyan), the invariant and the target (magenta), the variant and the target (yellow). The X-axis indicates the group number, and the Y-axis indicates the distance of the portion-hot representation. (d) The index alignment of the large parts along the 18 layers for the four images. The X-axis indicates the 5696 feature map indexes of the 18 layers, and the Y-axis indicates the four images. The marker size denotes the area ratio of a part in (b) and (d).}
	\label{fig7}
\end{figure*}

The part pathways for occluded images can reveal the cognition ability of the network. Similar results are shown in Fig. \ref{fig7}, with the part pathways of an original image, its crop-invariant version, its crop-variant version, and the target version. 

Both the perceptions of object movement and pattern completion are involved in implementing spatial and temporal integrative mechanisms. We conjecture that the degree of pathway matches between an object and its deformed version or between external stimuli and internal weights determines the ability to detect similarity, thereby enabling object recognition. Recognition of partially visible objects requires longer reaction times, and the neural representation of such objects is physiologically delayed by approximately 50ms with respect to the time required for fully visible objects \citep{a9}. The delay could be ascribed to the conduction of signals along the pathway \citep{a48} in addition to recurrent computations of lateral connections \citep{a9}.

\section{Discussion and Conclusion}
The stimulus-response methods do not reveal why a network is able to be highly sensitive to an object's shape, and the strong correlation between units and artificially segmented objects is still based on statistical calculations rather than on the inherent operation of a network. 
We develop an understanding of how a network spontaneously perceives regions at specific locations or part shapes and an algorithm to analyze the roles of the diffusion pathways of an image within classification networks. In artificial neural networks (ANNs), the location of an individual pixel is actually passed to each feature map of every layer through the diffusion kernels to distinguish and segment the region of an image that matches an important feature map for classification. We have seen that the part pathways of images of the same category are coincidental and the portion-hot representations are significantly different between categories.

We conclude that a systematic analysis of the diffusion pathway of an image can yield insight into the inner workings of the deep networks. By observing and manipulating the part pathways, it is possible to understand the latent position conduction mechanism in ANNs during the process of object recognition and help humans explore the behavior of BNNs. Although our algorithm for extracting the diffusion pathway might be complex and energy-consuming, the conduction of such pathways in BNNs may be simpler and more efficient. This is because BNNs rely on intermittent rather than continuous sampling from the outside world.

\section*{Acknowledgements}
This work was supported in part by the National Natural Science Foundation of China (Nos. 61976127, 62276156), Shandong Provincial Natural Science Foundation (Nos. ZR2021LZL012, ZR2022LZH004, ZR2021MF048), and the Taishan Scholar Program of Shandong Province of China (No. tsqn202306150).







\printcredits

\bibliographystyle{unsrt}

\bibliography{mybibfile}

\bio{}
\endbio

\endbio

\end{document}